\documentclass[runningheads]{llncs}

\usepackage{eccv}

\usepackage[dvipsnames]{xcolor}

\definecolor{green1}{RGB}{119, 204, 0}
\definecolor{dianqing}{HTML}{177cb0}
\definecolor{zangqing}{HTML}{2e4e7e}
\definecolor{juhong}{HTML}{ff7500}

\newcommand{\M}{${M^2D}$NeRF}
\newcommand{\fullM}{Multi-Modal Decomposition NeRF}

\newcommand{\lossmms}{$\mathcal{L}_{mms}$\xspace}
\newcommand{\losscl}{$\mathcal{L}_{cl}$\xspace}
\newcommand{\lossclv}{$\mathcal{L}_{cl}^v$\xspace}

\expandafter\def\expandafter\normalsize\expandafter{%
    \normalsize%
    \setlength\abovedisplayskip{2pt}%
    \setlength\belowdisplayskip{2pt}%
    \setlength\abovedisplayshortskip{0pt}%
    \setlength\belowdisplayshortskip{0pt}%
}

\usepackage{eccvabbrv}

\usepackage{graphicx}
\usepackage{booktabs}

\usepackage{pifont}
\usepackage{multirow}

\usepackage[accsupp]{axessibility}  

\usepackage[pagebackref,breaklinks,colorlinks,citecolor=eccvblue]{hyperref}

\usepackage{orcidlink}

\begin{document}

\title{\M: Multi-Modal Decomposition NeRF with 3D Feature Fields}


\author{Ning Wang\inst{1}\thanks{This project was undertaken by Ning Wang as a visiting student at Simon Fraser University, funded by the China Scholarship Council (CSC).} \and
Lefei Zhang\inst{1}\thanks{Corresponding author} \and
Angel X Chang\inst{2}}

\authorrunning{F.~Author et al.}

\institute{Wuhan University \and Simon Fraser University
}

\maketitle

\newcommand{\figfirstpagefigure}{
\vspace{-0.4cm}
  \begin{center}
  \captionsetup{type=figure}
  \includegraphics[width=0.87\textwidth]{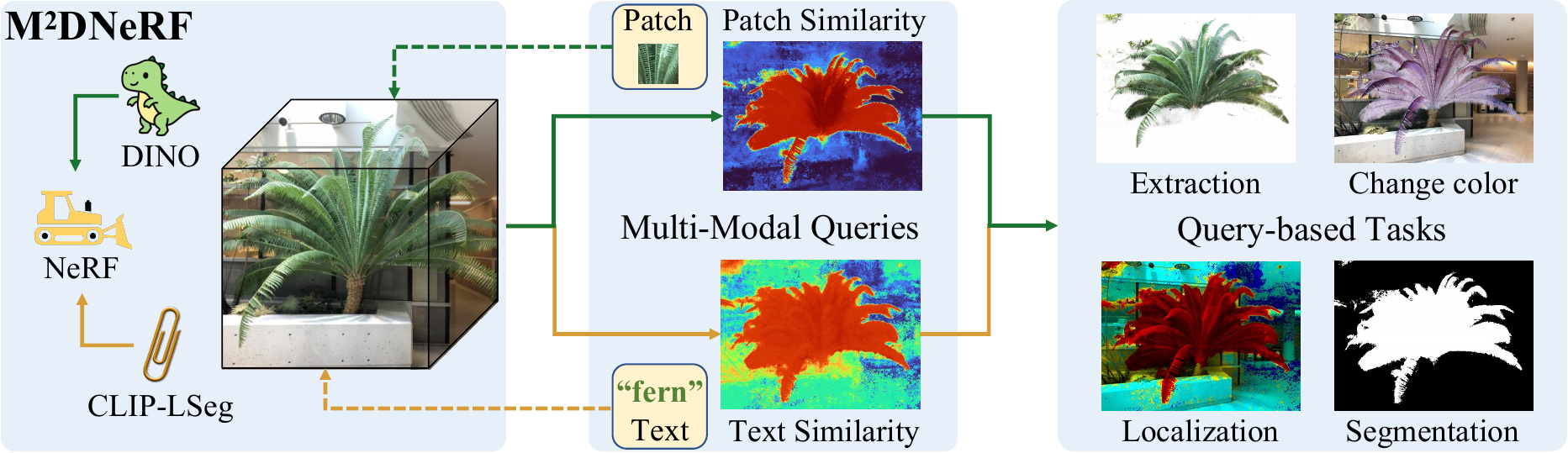}
  \captionof{figure}{ \textbf{{\fullM} ({\M}).} Using DINO and CLIP-LSeg as teacher models for feature distillation, our {\M} facilitates multi-modal queries for 3D scene decomposition. By establishing explicit relationships within multi-modal feature fields, our {\M} supports tasks with both text and patch-based queries and exhibits sharper and more accurate object identification.}
  \label{fig:teaser}
\end{center}
  \vspace{-0.5cm}
}

\figfirstpagefigure

\begin{abstract}
Neural fields (NeRF) have emerged as a promising approach for representing continuous 3D scenes. 
Nevertheless, the lack of semantic encoding in NeRFs poses a significant challenge for scene decomposition.
To address this challenge, we present a single model, {\fullM} ({\M}), that is capable of both text-based and visual patch-based edits. 
Specifically, we use multi-modal feature distillation to integrate teacher features from pretrained visual and language models into 3D semantic feature volumes, thereby facilitating consistent 3D editing. 
To enforce consistency between the visual and language features in our 3D feature volumes, we introduce a multi-modal similarity constraint. 
We also introduce a patch-based joint contrastive loss that helps to encourage object-regions to coalesce in the 3D feature space, resulting in more precise boundaries.
Experiments on various real-world scenes show superior performance in 3D scene decomposition tasks compared to prior NeRF-based methods. 
\end{abstract}


\section{Introduction}
\label{sec:intro}

Scene understanding and editing are crucial for computer vision and computer graphics. They are fundamental for the development of intelligent robots, AR/VR, and other applications. 
Given multi-view images of a scene, the goal of scene editing is to allow intuitive 3D edits, such as removing the slippers from the floor. 
Before any edits, it is essential to precisely localize and \emph{decompose} the target region from surrounding area.
Neural Radiance Fields (NeRF)~\cite{nerf2020mildenhall} and its variants~\cite{dvgo2022sun, plenoxels2022fridovich_and_yu,instantngp2022muller, mipnerf3602022barron, 3dgs2023kerbl} show promising results in novel view synthesis and high-quality 3D reconstruction from multi-view images. 
However, 3D scene decomposition remains challenging for NeRF-based representations as the captured scenes lack object-level awareness and semantic information for scene interaction.

Recent work has started to explore the ability of NeRFs for scene understanding and decomposition.
SemanticNeRF~\cite{semanticnerf2021zhi}, ObjectNeRF~\cite{objectnerf2021yang}, and Volumetric Disentanglement~\cite{volumetricdisentanglement2022benaim} decompose scenes by using object-level supervisory signals such as 2D masks or semantic/instance labels. 
However, their reliance on object-level supervision requires substantial effort to acquire accurate 3D or multi-view annotation.
These methods can also suffer from 3D space ambiguity caused by inaccurate annotation. 
By using synthetic scenes, OSF~\cite{osf2020guo}, uORF~\cite{uorf2022yu}, and DM-NeRF~\cite{dmnerf2023wang} eliminated the need for manual annotation. 
However, the domain gap between synthetic and real-world data poses a significant challenge, thereby restricting applicability to real-world scenes.
SA3D~\cite{sa3d2023cen} and SA-GS~\cite{sags2024hu} use SAM~\cite{sam2023kirillov} as mask extractor to generate efficient multi-view masks. Although they support real-world scenes, the extracted masks may lack 3D consistency and restrict to a specific object in single model.

Feature distillation enables decomposition of scenes
without the need for 3D semantic annotation and supports any object edits in a single model. N3F~\cite{n3f2022tschernezki} and DFF~\cite{dffs2022kobayashi} used feature distillation to propagate existing semantic knowledge from pretrained 2D models to a 3D-consistent semantic feature volume. 
They supervise the feature field with a pretrained 2D zero-shot encoder as the teacher network. 
Then the feature spaces capture the semantic properties of regions and make it possible to decompose scenes by text or image query. 
ISRF~\cite{isrf2023goel} uses a similar feature distillation strategy, which enables fine-grained interactive segmentation with user-provided strokes. 
These methods demonstrate the effectiveness of the feature distillation strategy in endowing NeRF models with semantics, promoting scene understanding and decomposition capability, and obviating the cost of human annotation. 

However, these methods encounter a common challenge, namely, the inability to precisely isolate designated objects, due to semantic ambiguity at object boundaries. 
These methods rely on feature similarity for scene decomposition. 
During decomposition, N3F~\cite{n3f2022tschernezki} and DFF~\cite{dffs2022kobayashi} calculate the feature similarity between user-provided cues and distilled feature fields to pick out the most similar regions to the input query. 
However, the boundary of the specified object is usually affected by contextual information, which tends to decrease its similarity to the user-provided cues. 
Although ISRF~\cite{isrf2023goel} attempted to solve this problem by region growing, it still struggles with accurate separation. The model tends to be confused in the boundary regions. 
Moreover, N3F and ISRF only support visual queries, such as image patches and strokes. 
The DFF model enables NeRF to decompose a specific object with a text query or an image patch query, but it doesn't support multi-modal queries with a single model.

In this work, we propose a novel {\fullM} ({\M}) to address the aforementioned challenges. Specifically, we adopt a multi-modal feature distillation strategy to accommodate both text and image patch queries. Our {\M} extends the NeRF model by introducing both a visual feature branch and a language feature branch.  We optimize the 3D visual and language feature volumes through the distillation of 2D DINO~\cite{dino2021caron} and 2D LSeg~\cite{lseg2022li} features, respectively. 
Although this adaptation enables multi-modal NeRF decomposition, the relationship between two feature volumes remains relatively weak. 
To address this limitation, we further devise a multi-modal similarity constraint, aimed at establishing a more explicit relationship between the two 3D feature volumes. 
The joint learning of visual and language features enhances the semantic understanding capabilities of NeRF models. 
We also employ another patch-based joint contrastive learning scheme to mitigate boundary-related challenges. 
This scheme leverages the 3D features from two semantic feature volumes to establish relationships pertaining to appearance and geometry. 
These two semantic feature volumes are subsequently optimized by pulling/pushing positive/negative pairs. 
The utilization of this joint contrastive learning scheme produces object features with distinct boundaries.

To summarize, the core contributions of our work are:
\begin{itemize}
    \item We introduce the {\fullM} ({\M}) model for 3D scene editing. By leveraging DINO and LSeg as teacher models and utilizing the NeRF model as the student network, we employ a multi-modal feature distillation strategy to create two 3D semantic feature volumes. These volumes offer versatile support for 3D scene editing, accommodating queries from both image patches and text prompts.
    \item We innovate with a multi-modal similarity constraint, emphasizing the cosine similarity between two 3D feature volumes. This constraint fosters an explicit relationship between these two feature volumes, enabling the NeRF model to harness both visual and language features to enhance its performance.
    \item We employ a novel joint contrastive learning scheme, which enhances the distinction between different object features. This technique leads to more pronounced differences and, as a result, facilitates more precise boundary.
\end{itemize}
\section{Related Work}
\label{sec:related}
\subsection{Neural Rendering and Editing}


3D scene editing is a complex task, requiring a considerable degree of semantic knowledge to be encoded in the 3D representation. Even though explicit 3D representations, such as polygonal mesh, allow for editing, it requires expertise in 3D modeling software to conduct manual editing~\cite{objmani2014kholgade}. These traditional methods are also computationally burdensome and impractical when applied to real-world scenes with intricate and complex objects~\cite{ftetwild2020hu}.

NeRF~\cite{nerf2020mildenhall} has demonstrated remarkable performance in 3D scene representation based on differentiable volume neural rendering. Various works have studied the potential improvements of different aspects of NeRF, like the novel view synthesis quality~\cite{mipnerf2021barron, zipnerf2023barron}, training and inference speed~\cite{dvgo2022sun, plenoxels2022fridovich_and_yu, gaussiansplatting2023kerbl}, few-shot setting~\cite{sparsenerf2023wang, freenerf2023yang}, and compression~\cite{vqad2022takikawa, vqrf2023li}. NeRF has also been widely used for 3D-aware image generation~\cite{giraffe2021niemeyer2021}, text-to-3D generation~\cite{magic3d2023lin}, depth estimation~\cite{nerfingmvs2021wei}, pose estimation~\cite{nopenerf2023bian}, editing~\cite{clipnerf2022wang, nerfediting2022yuan, gaussianeditor2024fang, gaussianeditor2023chen}, and many other tasks.

Neural rendering has spurred an exploration into implicit and hybrid representations, offering various approaches for 3D editing, such as changing global appearance~\cite{neuralpil2021boss, palettenerf2023kuang}, intrinsic decomposition \cite{intrinsicnerf2023ye, i2sdf2023zhu}, per-object decomposition~\cite{objectnerf2021yang, objectsdf2022wu}, geometry and texture editing~\cite{neumesh2022yang, nerfediting2022yuan, sine2023bao}, 3D inpainting~\cite{spin2023mirzaei, removingnerf2023weder}, and others~\cite{instructnerf2nerf2023haque, clipnerf2022wang, nsg2021ost}. In this paper, our framework is developed based on Instant-NGP \cite{instantngp2022muller} and improves scene decomposition without human annotation. 

\subsection{NeRF-based Scene Decomposition}

\noindent \textbf{Label-based Decomposition.}
Various techniques have been proposed to address the challenges of 3D scene decomposition. DM-NeRF~\cite{dmnerf2023wang} learns object code for each object with 2D object labels and devises an inverse query algorithm to edit specified object shapes. Object-NeRF \cite{objectnerf2021yang} designs a two-pathway architecture, consisting of the scene branch and the object branch, for indoor object-composite scenes. Object-NeRF supports object movement, removal, or duplication. ObjectSDF~\cite{objectsdf2022wu} models the object-level geometry directly, allowing for the concurrent learning of geometry and semantics in complex scenes by capturing the Signed Distance Function (SDF) of each object. Benaim \etal~\cite{volumetricdisentanglement2022benaim} also disentangle foreground and background. However, these works predominantly work with synthetic scenes or require costly annotations for 2D instance segmentation, limiting their applicability to real-world scenes. 
NVOS~\cite{nvos2022ren} selects objects with the use of positive and negative user scribbles in one or more views, reducing the substantial cost associated with precise annotation. However, NVOS struggles to produce faithful segmentation while incurring significant performance overhead.
SA3D~\cite{sa3d2023cen} and SA-GS~\cite{sags2024hu} reduce the cost of annotation by generating masks with SAM~~\cite{sam2023kirillov}. However, optimization is required for every editing request, limiting the efficiency of scene editing.

\noindent \textbf{Distillation-based Decomposition.}
Recent works have used feature distillation to take pretrained image features and propagate them to neural volumetric representations.  This strategy helps to eliminate the need for annotations and enable scene decomposition within real-world and complex scenes.
DFF \cite{dffs2022kobayashi} distills the knowledge of pretrained 2D image feature extractors (CLIP-LSeg~\cite{lseg2022li} or DINO~\cite{dino2021caron}) into a 3D feature field, allowing for query-based local editing of NeRFs with textual prompts or patches used as the segmentation cues.
A concurrent work N3F \cite{n3f2022tschernezki} also explores the same training strategy with different teacher models (MoCo-v3 \cite{mocov32021chen}, DeiT \cite{deit2021touvron}, and DINO \cite{dino2021caron}) and NeuralDiff-based student model \cite{neuraldiff2021tschernezki}. N3F uses user-provided patches to semantically edit the radiance field. ISRF \cite{isrf2023goel}, building upon the hybrid implicit-explicit representation framework TensoRF \cite{tensorf2022chen}, harnesses user strokes as input for interactive editing. ISRF enhances the 3D segmentation through nearest-neighbor feature matching following feature distillation.

Feature distillation is also used in LERF \cite{lerf2023kerr}, FeatureNeRF \cite{featurenerf2023ye}, SparseNeRF \cite{sparsenerf2023wang}, and SUDS \cite{suds2023turki} to improve the ability of scene understanding, thus further enhances the performance on text-guided localization, generalization, few-shot novel view synthesis, and large-scale scene reconstruction, respectively.
These methods demonstrate the effectiveness of feature distillation to endow NeRF models with rich semantics and promote understanding capability. Moreover, this feature distillation strategy provides a notable benefit in terms of mitigating the demand for labor-intensive annotation efforts.
Our work builds upon this line of work and focuses on how to effectively combine information from several different teacher models across multi-modalities.  



\begin{figure*}[t]
    \centering
    \includegraphics[width=\textwidth]{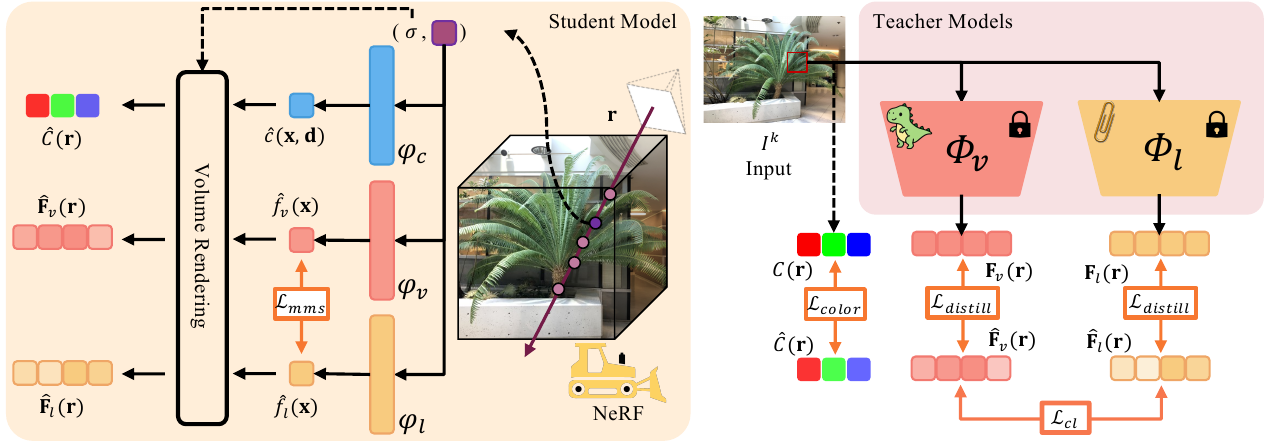}
    \caption{{\M} framework. \emph{Left}:  We expand the NeRF backbone by introducing two additional branches $\varphi_{v}$ and $\varphi_{l}$ ($\varphi_{c}$ denotes the original color branch). Given a 3D sample point $\mathbf{x}$, the three branches produces the volumetric color $\hat{c}(\mathbf{x},\mathbf{d})$ or feature $\hat{f}(\mathbf{x})$ (consisting of the visual $\hat{f}_v(\mathbf{x})$ and language $\hat{f}_l(\mathbf{x})$ features). By considering the volume density $\sigma$, we get the rendered color $\hat{C}(\mathbf{r})$ and visual $\hat{\mathbf{F}}_v(\mathbf{r})$ and language $\hat{\mathbf{F}}_l(\mathbf{r})$ features for each ray. \emph{Right}: We extract per-pixel multi-modal features $\mathbf{F}_{v}(\mathbf{r})$ and $\mathbf{F}_{l}(\mathbf{r})$ using models $\Phi_{v}$ and $\Phi_{l}$ (visual model DINO \cite{dino2021caron} and language model CLIP-LSeg \cite{lseg2022li} in our work). These act as 2D teacher models that supervise the student feature fields via distillation loss ($\mathcal{L}_{distill}$). We build both 3D (multi-modal similarity $\mathcal{L}_{mms}$) and patch-level 2D (joint contrastive learning scheme $\mathcal{L}_{cl}$) relationships between multi-modal features for better scene understanding.}
    \label{fig:method}
    \vspace{-0.5cm}
\end{figure*}

\section{Method}
\label{sec:method}

\Cref{fig:method} shows the our {\M} framework. We first review NeRF (\cref{sec:preliminaries}), before providing an overview of our multi-modal feature fields (\cref{sec:featurefields}).  We then describe how we optimize the different branches of our model (\cref{sec:loss}). Finally, we describe how the optimized multi-modal feature fields can be used for decomposition (\cref{sec:query}).

\subsection{Preliminaries: Neural Radiance Fields (NeRF)}
\label{sec:preliminaries}

NeRF \cite{nerf2020mildenhall} is a compact representation of 3D scenes through differentiable volume rendering. NeRF employs a 5D mapping function $\mathcal{F}_\theta:(\mathbf{x},\mathbf{d})\rightarrow(\mathbf{c},\sigma)$ to encode a 3D scene, which is parameterized by an MLP with learnable parameters $\theta$ and projects a spatial location $\mathbf{x} \in \mathbb{R}^3$ and a ray direction $\mathbf{d} \in \mathbb{S}^2$ to the corresponding view-dependent color $\mathbf{c} \in \mathbb{R}^3$ with a volume density $\sigma$.
Rendering an image given its corresponding camera pose is done by casting a camera ray $\mathbf{r}(t)=\mathbf{o}+t\mathbf{d}$ through each pixel, where $\mathbf{o} \in \mathbb{R}^3$ is the camera center/origin and $\mathbf{d} \in \mathbb{R}^3$ is the view direction. 

The color of the camera ray $\mathbf{r}$ then is estimated using numerical quadrature \cite{quadrature1995max}: 
\begin{equation}
\hat{\mathbf{C}}(\mathbf{r})=\sum_i T_i \alpha_i \mathbf{c}_i,
\label{equ:renderrgb}
\end{equation}
where $T_i =\mathrm{exp}(-\sum_{j=1}^{i-1}\sigma_j\delta_j)$, $\alpha_i =1-\mathrm{exp}(-\sigma_i\delta_i)$, $\delta_i =t_i-t_{i-1}$.
Here, $T_i$ denotes accumulated transmittance between points $\mathbf{r}(t_{i-1})$ and $\mathbf{r}(t_{i})$, $\alpha_i$ is the opacity, and $\delta_i$ denotes the step size.
The process of volume rendering is differentiable, which allows NeRF to be optimized with only 2D image supervision. The parameter $\theta$ is optimized by gradient descent with a Mean Square Error (MSE) loss between the estimated pixel color and the actual color:
\begin{equation}
\label{equ:colorloss}
\mathcal{L}_{color}=\sum_{\mathbf{r}} \|\hat{\mathbf{C}}(\mathbf{r}) - \mathbf{C}(\mathbf{r})\|_2.
\end{equation}


After optimization, the MLPs encode scenes as the implicit radiance field. Recently, pure voxel grids \cite{plenoxels2022fridovich_and_yu} and hybrid of explicit grids and MLPs \cite{instantngp2022muller} have been verified to be effective alternatives to implicit representation.
In this work, we use Instant NGP as backbone, which is much more efficient to train than the vanilla NeRF model. 


\subsection{Multi-modal Feature Fields}
\label{sec:featurefields}

To support scene decomposition, it is crucial for the 3D scene representation to capture semantics. This is particularly challenging for implicit 3D representations. In this work, we draw inspiration from earlier research~\cite{dffs2022kobayashi, n3f2022tschernezki} to use feature distillation to imbue our NeRF model with semantic knowledge and make it more suitable for semantic-aware decomposition. Although prior work has advanced the field of 3D scene understanding and editing, there is still significant room for improvement, particularly in handling multi-modal queries and addressing the precise delineation of object boundaries. 
We design NeRF with \emph{multi-modal feature fields} to enhance scene understanding and editing capabilities of implicit representations. 

Our approach incorporates two pretrained teacher models and one student model. Specifically, we integrate a visual foundation model (DINO \cite{dino2021caron}) and a language foundation model (LSeg \cite{lseg2022li}). 
The extracted teacher features from the visual and language models are denoted as $\mathbf{F}_{v}$ and $\mathbf{F}_{l}$, respectively. 
To construct the 3D multi-modal feature fields, we extend the backbone by adding two additional feature branches $\varphi_{v}$ and $\varphi_{l}$. Each branch is implemented through a simple MLP layer, producing a feature vector $\hat{\mathbf{f}}_{m}(\mathbf{x})$, where $m \in {\{v,l\}}$ indicates the feature type. For a given camera ray $\mathbf{r}$, the corresponding feature $\hat{\mathbf{F}}_{m}(\mathbf{r})$ is calculated via volume rendering similar to \cref{equ:renderrgb}. Subsequently, the feature fields are distilled using the extracted multi-modal features to construct the respective feature fields.

\begin{figure*}[t]
  \centering
   \includegraphics[width=\linewidth]{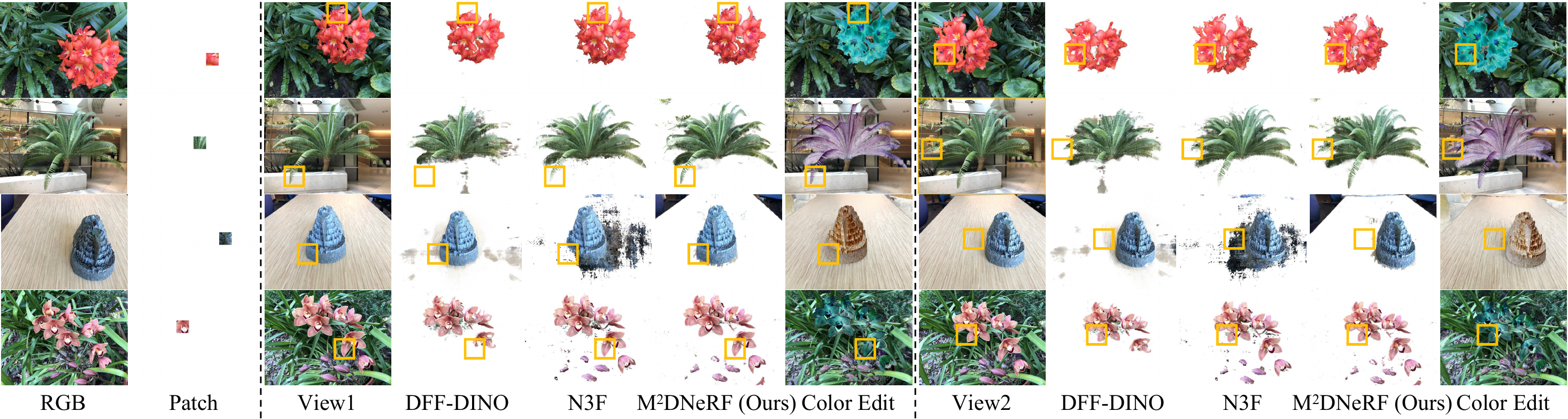}
   \vspace{-0.6cm}
   \caption{Patch query-based 3D extraction and color edit. We compare with DFF-DINO and N3F showing two views that highlight the multi-view consistency in 3D object extraction. We recommend zooming in and focusing on the yellow boxes to inspect intricate boundaries and fine details. We can then edit these extracted 3D regions, like changing the color (last column for each view).}
   \label{fig:patch}
   \vspace{-0.6cm}
\end{figure*}

\subsection{Loss Functions}
\label{sec:loss}

To optimize our multi-modal feature fields, we enhance the base NeRF model with three different loss terms consisting of 1) a \emph{distillation loss} $\mathcal{L}_{distill}$ that encourages the 3D visual and language features to match features from the pretrained foundation models, 2) a \emph{multi-modal similarity} $\mathcal{L}_{mms}$ to encourage the visual and language features to match each other, and 3) a patch-based \emph{joint contrastive learning loss} $\mathcal{L}_{cl}$ that encourages patches of the 3D volume to have similar features if features from the corresponding patches in the rendered 2D images have similar features. As described earlier, the distillation loss follows prior work that uses feature distillation to encode semantics into NeRF models. Here, we extend it to multi-modal feature fields that combine vision and language. However, due to differences in the vision and language features, we propose the two additional losses to help align the vision and language features, and to encourage regions with similar semantics to have similar features so that the boundary is sharper and more precise. Each of the loss terms is described in more detail below.

\noindent \textbf{Distillation Loss.}
The distillation loss $\mathcal{L}_{distill}$ penalizes the difference between the \emph{rendered} feature $\hat{\mathbf{F}}_{m}(\mathbf{r})$ and the \emph{extracted} feature $\mathbf{F}_{m}(\mathbf{r})$ from teacher models,
\begin{equation}
    \mathcal{L}_{distill} = \sum_{\mathbf{r}} \|\hat{\mathbf{F}}_{m}(\mathbf{r}) - \mathbf{F}_{m}(\mathbf{r})\|_2.
\end{equation}

Since the two feature branches $\varphi_{v}$ and $\varphi_{l}$ share the same architectural backbone, the simultaneous incorporation of DINO and LSeg features helps to capture the multi-modal semantics of 3D scenes. While both $\varphi_{v}$ and $\varphi_{l}$ can independently conduct feature matching based on the provided query to decompose the most relevant regions, the regions obtained from text queries are notably less accurate than those obtained from image patches. Additionally, there is a tendency for ambiguity in defining object boundaries in both cases. 
Hence, we propose two solutions to address these issues, both grounded in establishing explicit multi-modal relationships.

\noindent \textbf{Multi-Modal Similarity.} The first solution is to constrain the similarity between visual features and language features. As the 2D features obtained from LSeg tend to be more ambiguous compared to those from DINO, the distilled 3D language feature field appears coarser compared to the 3D visual feature field. To enhance text-guided decomposition, one straightforward approach is to encourage similarity between the multi-modal feature fields. Therefore, we introduce the multi-modal similarity loss for our multi-modal feature fields in the following manner:
\begin{equation}
    \mathcal{L}_{mms} = 1 - \mathbf{cos}(\hat{\mathbf{f}}_{v}(\mathbf{x}), \hat{\mathbf{f}}_{l}(\mathbf{x})),
\end{equation}
where $\mathbf{cos}()$ denotes the cosine similarity function.

\begin{figure}[t]
  \centering
   \includegraphics[width=0.9\linewidth]{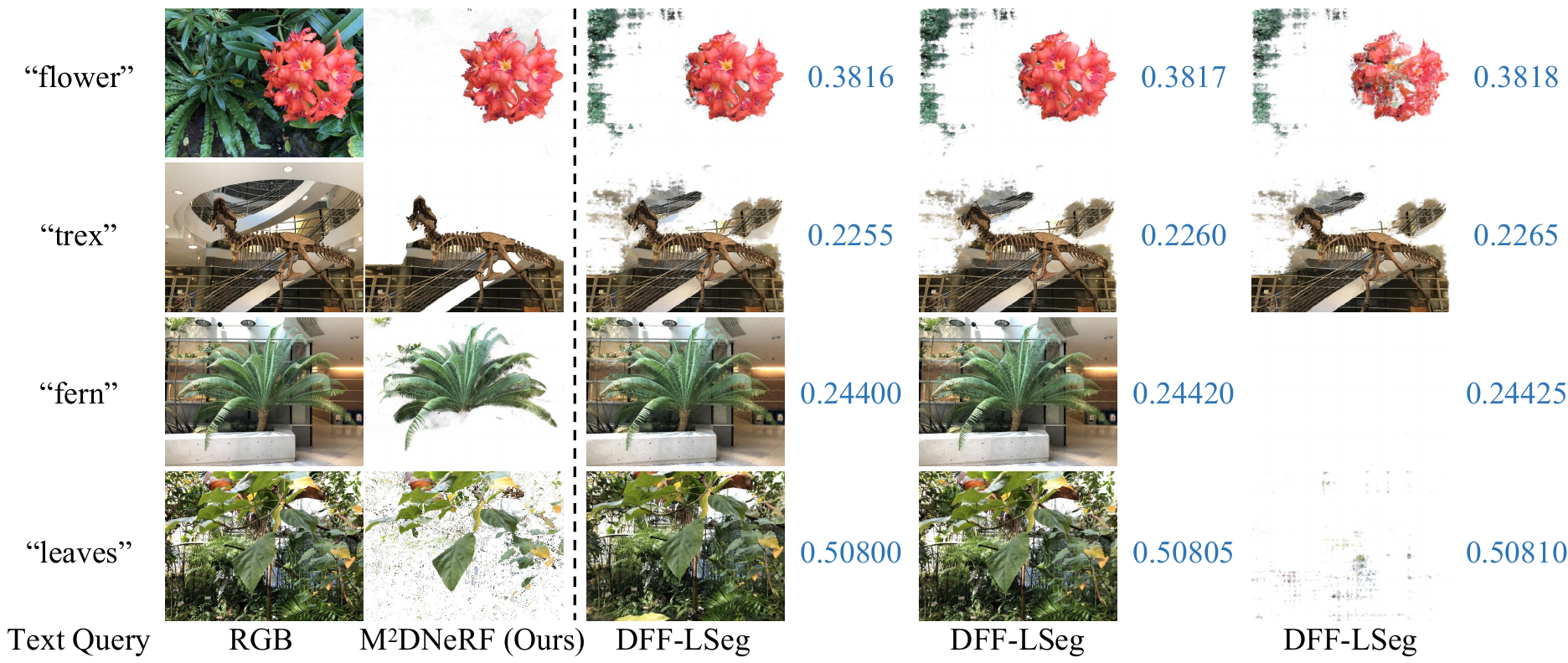}
   \vspace{-0.3cm}
   \caption{Qualitative results with text query. Our {\M} effectively extracts corresponding objects based on text queries (left column) with a fixed threshold, showcasing superior scene understanding. In contrast, DFF-LSeg struggles to find a suitable threshold for precise object extraction. With just tiny changes to the threshold (see blue numbers for thresholds used for DFF-LSeg), DFF-LSeg can end up extracting almost the entire image or nothing (see last rows).}
   \label{fig:text}
   \vspace{-0.7cm}
 \end{figure}


\noindent \textbf{Joint Contrastive Learning.} Drawing inspiration from \cite{stego2021hamilton} and \cite{nerfsos2022fan}, we propose a \emph{patch-based} joint contrastive learning scheme as the second solution to establish relationships between \emph{patches} for our multi-modal feature fields. The use of contrastive losses at the patch-level helps to drive semantically similar regions to have more similar features, and features for semantically different regions to have lower similarity, thus allowing for sharper and more distinct boundaries. 
We build patch-level relationships for (appearance, multi-modal features) and (geometry, multi-modal features).
Since both appearance and geometry have correlation patterns that are consistent with semantic labels, by ensuring that the 3D visual feature field and language feature field align with the same appearance and geometry correspondences, our multi-modal feature fields will also have consistent distributions and semantics.
 
\begin{table*}[t]
\centering
\caption{Quantitative results for rendering and segmentation. 
We report the rendering metrics (PSNR, SSIM, LPIPS) by averaging over eight commonly used scenes within the LLFF dataset and segmentation metrics (IoU) for the \textit{\{Flower, Fortress\}} scenes only.  
For NeRF-SOS and Semantic-NeRF, the rendering metrics are over the \textit{\{Flower, Fortress\}} scenes only (indicated by *), as they rely on pretrained checkpoints or semantic labels which are only available for these two scenes. Note that Semantic-NeRF is a supervised method, and the results for segmentation are derived from NeRF-SOS.
}
\vspace{-0.3cm}
\begin{tabular}{lccccccc}
\toprule
Methods   & Feature   & PSNR$\uparrow$ & SSIM$\uparrow$ & LPIPS$\downarrow$ & IoU(BG)$\uparrow$ & IoU(FG)$\uparrow$ & mIoU$\uparrow$ \\
\midrule
NeRF-SOS* \cite{nerfsos2022fan}    &       & (\textbf{27.85})  & (0.9070)  & (0.0587)   & 97.95  & 91.41  & 94.68    \\
Semantic-NeRF*  \cite{semanticnerf2021zhi}& & (27.65)  & (0.8039)  & (0.1323)   & 98.53      & 94.45       & 96.49    \\
\midrule

DFF-Lseg \cite{dffs2022kobayashi} &  \multirow{3}{*}{Text}       & 25.12    & 0.8683    & 0.0964     & 65.95  & 47.86  & 56.90    \\
LERF \cite{lerf2023kerr}        &           & 14.49    & 0.5612    & 0.3252     & 81.19  & 49.39  & 65.29    \\
{\M} (Ours)     &                     & 27.80   &  \textbf{0.9139}   &  \textbf{0.0474}    & 98.17       & 92.94      & 95.55    \\
\midrule

DFF-DINO \cite{dffs2022kobayashi}  & \multirow{3}{*}{Patch}       & 25.21    & 0.8699    & 0.0927     & 95.40  & 80.10  & 87.75    \\
N3F \cite{n3f2022tschernezki}     &         & 27.63    & 0.9074    & 0.0614     & 93.94  & 76.17  & 85.05    \\
{\M} (Ours)  &  & 27.80   &  \textbf{0.9139}   &  \textbf{0.0474}   & \textbf{98.83 }      & \textbf{94.87 }      & \textbf{96.85}   \\
\bottomrule
\end{tabular}
\label{tab:segmentation}
\vspace{-0.7cm}
\end{table*}

In contrast to the first multi-modal similarity loss applied to the 3D feature fields, the contrastive scheme revolves around the selection of positive and negative pairs of patches. 
Therefore, this constraint is implemented for 2D patches or features. 
We use the rendered multi-modal features $\hat{\mathbf{F}}$ to build multi-modal semantic feature correspondence $\mathbf{S}_{hwh'w'}$ across views as follows (the subscript $m$ is omitted for brevity and clarity),
\begin{equation}
\mathbf{S}_{hwh'w'} = \sum_c \frac{\hat{\mathbf{F}}_{chw}}{|\hat{\mathbf{F}}_{hw}|} \frac{\hat{\mathbf{F}}'_{ch'w'}}{|\hat{\mathbf{F}}'_{h'w'}|}, 
\label{eqn:feat_correspondence}
\end{equation}
where $\hat{\mathbf{F}}_{chw}$ represents the rendered multi-modal features for the location $(h,w)$ with dimensions $C \times H \times W$, and the symbol $'$ indicates that the rendered feature is derived from random patches in different views, with $c$ iterating through the feature channel dimension.

Following the setup of Fan \etal~\cite{nerfsos2022fan}, the appearance correspondence is established by assessing the similarity between the visual features of two randomly selected patches. These visual features $f$ and $f'$ are obtained using DINO by feeding the rendered RGB patches.
\begin{equation}
F_{hwh'w'} = \sum_c \frac{f_{chw}}{|f_{hw}|} \frac{f'_{ch'w'}}{|f'_{h'w'}|}, 
\label{eqn:app_correspondence}
\end{equation}
Using the rendered depth $d$ as geometry cue, the geometry correspondence is calculated as follows,
\begin{equation}
G_{hwh'w'} = \sum_c  \frac{1}{\vert d_{chw} - d'_{ch'w'}\vert + \epsilon}.
\label{eqn:geo_correspondence}
\end{equation}

The appearance-feature correlation $\mathcal{C}_{app}$ and geometry-feature correlation $\mathcal{C}_{geo}$ are constructed as follows,
\begin{equation}
\mathcal{C}_{app}(\mathbf{r},b) = - \sum_{hwh'w'} (F_{hwh'w'} - b) \mathbf{S}_{hwh'w'},
\label{eqn-simple_coor} 
\end{equation}
\begin{equation}
\mathcal{C}_{geo}(\mathbf{r},b) = - \sum_{hwh'w'} (G_{hwh'w'} - b) \mathbf{S}_{hwh'w'},
\label{eqn-geo_coor}
\end{equation}
where $b$ is used to control the positive and negative pressure. Then we use contrastive learning to pull/push positive/negative pairs ($p$ and $n$ are used for positive pairs and negative pairs, respectively):
\begin{equation}
\mathcal{L_{\mathit{app}}} = \mathcal{C}_{app}(\mathbf{r}_{p}, b_{p}) + \mathcal{C}_{app}(\mathbf{r}_{n}, b_{n}),
\label{app-contrastive}
\end{equation}
\begin{equation}
\mathcal{L_{\mathit{geo}}} = \mathcal{C}_{geo}(\mathbf{r}_{p}, b_{p}) + \mathcal{C}_{geo}(\mathbf{r}_{n}, b_{n}).
\label{geo-contrastive}
\end{equation}

Our joint contrastive learning loss is then:
\begin{equation}
     \mathcal{L}_{cl} = \lambda_{v} (\lambda_{a} \mathcal{L}_{\mathit{app}}^{v} + \lambda_{g} \mathcal{L}_{\mathit{geo}}^{v}) + \lambda_{l}(\lambda_{a} \mathcal{L}_{\mathit{app}}^{l} + \lambda_{g} \mathcal{L}_{\mathit{geo}}^{l}),
\label{contrastive}
\end{equation}
where $\lambda_{v}$ and $\lambda_{l}$ indicate loss force between visual feature field and language feature field, and $\lambda_{a}$ and $\lambda_{g}$ denote the balancing weights for appearance coherence and geometry coherence, respectively.
For the constrastive loss, we select positive and negative pairs by identifying the highest and lowest values in each row of a cosine similarity matrix. This matrix is generated by summing the visual feature similarity and language feature similarity for each patch pair.

\noindent \textbf{Objective Function.}
We optimize {\M} by minimizing the combined loss:
\begin{equation}
    \mathcal{L}=\mathcal{L}_{color} + \lambda_{distll}\mathcal{L}_{distill} + \lambda_{mms}\mathcal{L}_{mms} + \lambda_{cl}\mathcal{L}_{cl},
\end{equation}
where $\lambda_{distill}$, $\lambda_{mms}$, and $\lambda_{cl}$ are balancing weights. $\mathcal{L}_{color}$ (Eq. \ref{equ:colorloss}) is the photometric loss as in NeRF \cite{nerf2020mildenhall}.

\subsection{Decomposition}
\label{sec:query}
Once we have our multi-modal feature fields, we can decompose the NeRF scenes based on the feature similarity of user query and 3D feature volumes. For the text queries, we extract the language features $q_{l}$ from the provided prompts using the text encoder of CLIP~\cite{clip2021radford}. In the case of image-patch queries, we extract the visual features $q_{v}$ using DINO~\cite{dino2021caron}. We then calculate the feature similarity $\mathcal{Q}_{m}(\mathbf{x})$ for each 3D point by comparing the query feature $q_{m}$ with the corresponding feature field, $\mathcal{Q}_{m}(\mathbf{x})= \mathbf{cos}(q_{m}, \hat{\mathbf{f}}_{m}(\mathbf{x}))$.
By applying a threshold to the feature similarity, i.e., $\|\mathcal{Q}_{m}\| \geq \tau_{m}$, we can extract out the queried object in the scene by setting the density $\sigma(\mathbf{x})$ to zero for all 3D points that do not correspond to that object. 
3D scene editing can be implemented by manipulating the localized parameters, like changing the color of the well-extracted 3D objects.

\begin{figure*}[t]
  \centering
   \includegraphics[width=0.95\linewidth]{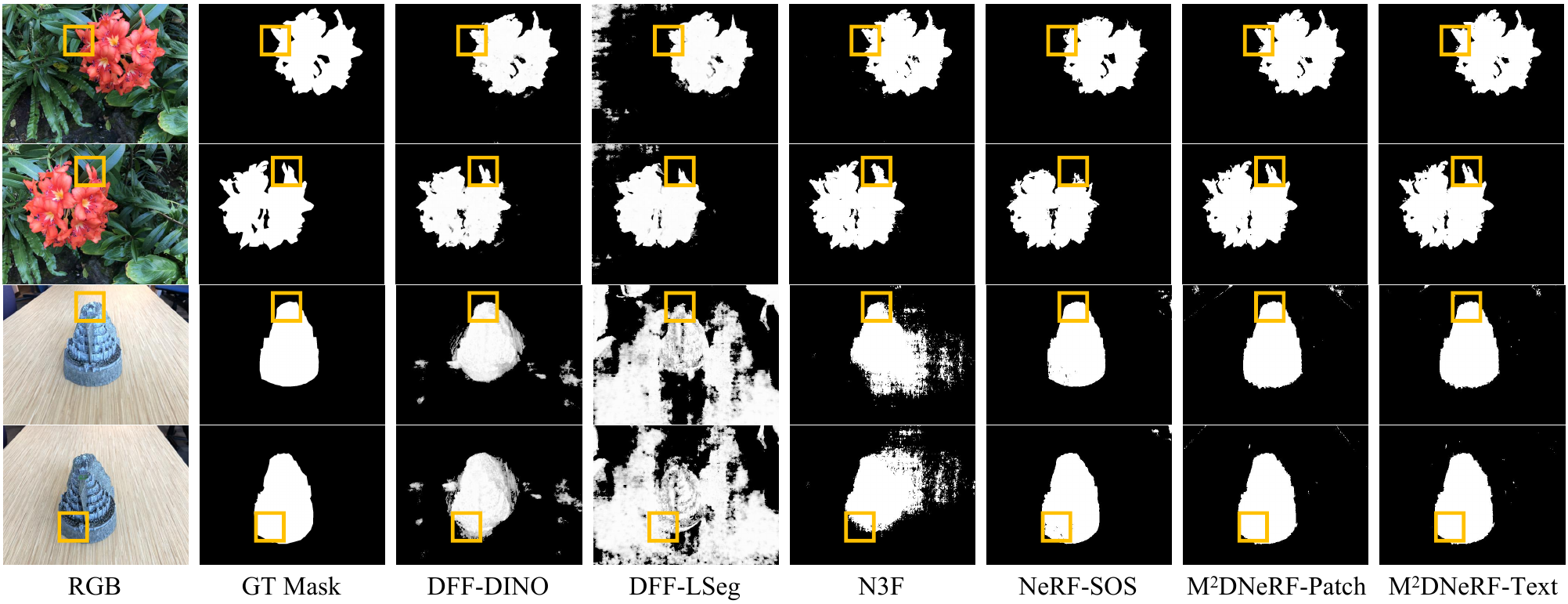}
   \vspace{-0.3cm}
   \caption{Segmentation results on \textit{\{Flower, Fortress\}} scenes. Compared with NeRF-SOS, our {\M} gets more complete objects. Compared with DFF and N3F, our method can greatly reduce the noise. Our results are more close to the ground-truth. We recommend zooming in and focusing on the yellow boxes to inspect details.}
   \label{fig:segmentation}
   \vspace{-0.3cm}
\end{figure*}

\begin{figure}[t]
  \centering
   \includegraphics[width=0.9\linewidth]{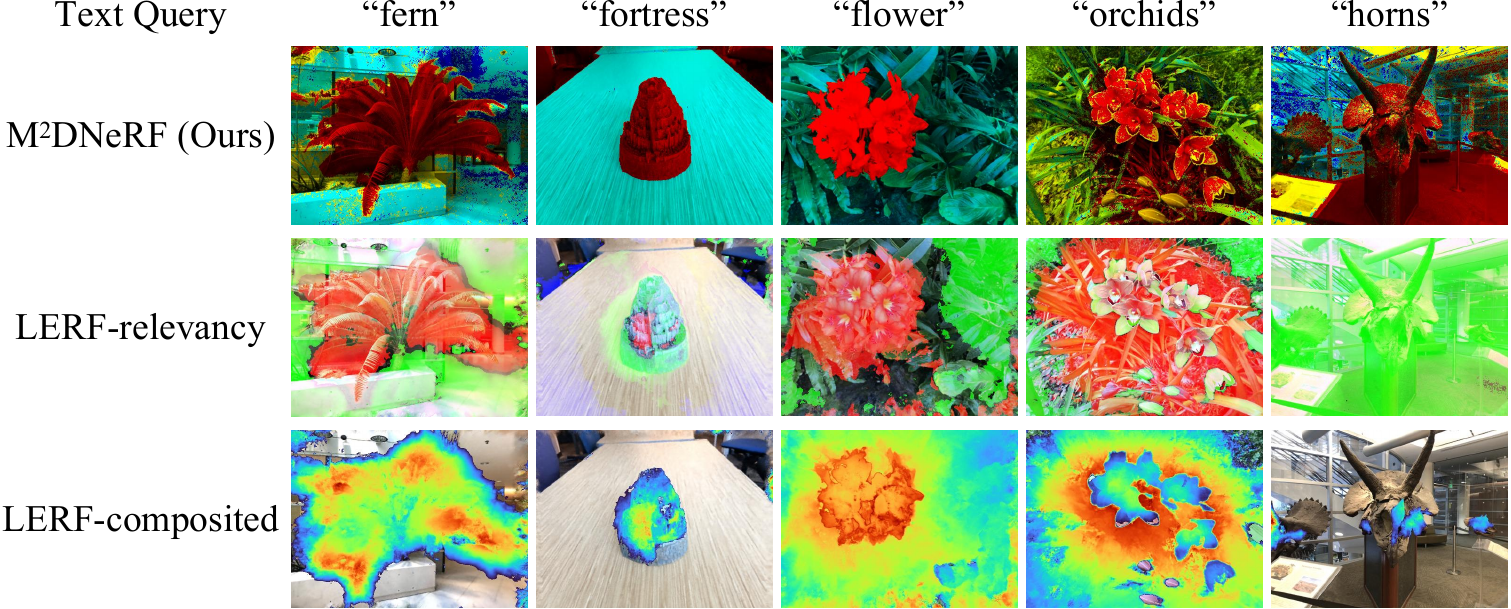}
   \vspace{-0.2cm}
   \caption{We compare our {\M} with LERF~\cite{lerf2023kerr} for localization. 
   For comparison, we superimpose relevancy results on RGB images ({\M} (Ours) and `LERF-relevancy'). We use a rainbow scale with blue being the least relevant, and red most relevant.  
   For LERF, we also show `LERF-composited' results rendered with NeRFStudio~\cite{nerfstudio} as originally shown in their paper. 
 }
   \label{fig:localization}
   \vspace{-0.5cm}
\end{figure}

\section{Experiments}
\label{sec:experiments}

In this section, we provide the experimental details for implementation and evaluation in \cref{sec:setup} and the comparative analysis of our {\M} against existing distillation-based scene decomposition approaches DFF~\cite{dffs2022kobayashi} and N3F~\cite{n3f2022tschernezki} in \cref{sec:decomposition}. The segmentation and localization performance are discussed in \cref{sec:seg} and \cref{sec:localization}. And the ablation study of {\M} is conducted in Sec. \ref{sec:ablation}.

\subsection{Experimental Setup}
\label{sec:setup}

\noindent \textbf{Datasets.}
To assess the performance of our method, we evaluate all methods on the representative benchmark \textbf{LLFF}~\cite{llff2019mildenhall}, which provides several forward-facing scenes that contain at least one common object among most views. For semantic segmentation evaluation, we make use of the binary masks provided by NeRF-SOS~\cite{nerfsos2022fan} for \textit{\{Flower, Fortress\}} scenes from the LLFF dataset. 

\noindent \textbf{Metrics.}
We provide quantitative evaluation (\cref{tab:segmentation}) for rendering and segmentation quality using the following metrics: (1) the rendering quality is evaluated by Peak Signal-to-Noise Ratio (PSNR),  Structural Similarity Index Measure (SSIM), and Learned Perceptual Image Patch Similarity (LPIPS)~\cite{lpips2018zhang}; and (2) the segmentation performance is measured by Intersection-over-Union (IoU) of the background (BG), foreground (FG), and the mean (mIoU).

\noindent \textbf{Implementation.}
We use Instant-NGP~\cite{instantngp2022muller} as the NeRF backbone, using the open-source PyTorch-Lightning implementation~\cite{nsr-pl}. We modify the architecture to include two additional feature branches for the language and vision features. Each of these branches has a single linear layer with a sigmoid activation function. During pretraining, we optimize the base model with photometric loss for 20k iterations. Then we continue training for 5k iterations to construct our multi-modal feature fields. For the 2D multi-modal teacher models, we consider two transformer-based feature extractors: DINO~\cite{dino2021caron} for visual features and LSeg~\cite{lseg2022li} for language features. We use DINO with the pretrained ViT-B/8 backbone. The extracted features are L2-normalized and reduced with PCA to 64 dimensions. When joint contrastive learning is used, first distill the multi-modal features with the entire network, and then we freeze the backbone and the RGB branch to prioritize optimization of the multi-modal features. The patch size for {\losscl} is fixed and set to 8$\times$8 for all experiments. The loss weights $\lambda_{distill}, \lambda_{mms}, \lambda_{cl}, \lambda_{v}, \lambda_{l}, \lambda_{a}, \lambda_{g}$ are set to 0.1, 0.01, 0.01, 1, 1, 0.001, and 0.01.
Our method is trained for 30k iterations in total, taking less than half an hour on a single NVIDIA RTX A5000 GPU with 24GB memory.

\subsection{Decomposition}
\label{sec:decomposition}

\noindent \textbf{Vision-based Decomposition.}
Given image patch queries, we present the vision-based results for DFF-DINO~\cite{dffs2022kobayashi}, N3F~\cite{n3f2022tschernezki}, and our {\M} in \cref{fig:patch}.
The query feature $q_{v}$ is calculated by averaging the extracted features of all patch pixels.  We then identify and decompose specific 3D regions by grouping feature similarities and selecting the corresponding parts based on the editing types, for instance, extraction and deletion choose opposite regions. \cref{fig:patch} shows that both DFF-DINO and N3F struggle to achieve a clear scene decomposition with well-defined boundaries or complete objects. In the \textit{Fern} scene (second row), both DFF-DINO and N3F neglect smaller leaves instead of extracting the entire fern as intended. 
In contrast, our {\M} excels in segmenting the specific 3D region. Despite our approach demonstrating an enhanced ability in boundary delineation when using image patch-based queries, it occasionally associates with incorrect regions (third row).
Clean object extraction enables accurate editing of the corresponding areas in 3D. We showcase the ability of our model to change the color (see last column for each view).

\noindent \textbf{Language-based Decomposition.}
For text-based queries, we compare the object extraction results obtained using DFF-LSeg~\cite{dffs2022kobayashi} and our {\M} (see \cref{fig:text}). For a given text query, we calculate the language feature by feeding the text query into the pretrained CLIP~\cite{clip2021radford}. In the original setting \cite{dffs2022kobayashi}, DFF-LSeg needs both positive (target) and negative texts as queries. Additionally, determining the appropriate threshold for feature selection poses a considerable challenge. Notably, our {\M} stands out as the top performer in text-guided scene decomposition, despite only requiring a single query for the target object. This superior performance is achieved by matching the multi-modal features. The language feature field shows improved boundary recognition capabilities by aligning its distribution with that of the visual feature field, which captures finer visually nuanced details than the language feature.


\begin{table*}[t]
\centering
\caption{Ablation study of multi-modal similarity loss and joint contrastive learning scheme with six different settings. The evaluation encompasses both DINO-feature based segmentation and LSeg-feature based segmentation.}
\vspace{-0.3cm}
\begin{tabular}{cccccccccc}
\toprule
\multicolumn{4}{c}{Flower Scene}   & \multicolumn{3}{c}{DINO}   & \multicolumn{3}{c}{LSeg} \\
\cmidrule(l){1-4} \cmidrule(l){5-7} \cmidrule(l){8-10}
Settings    &$\mathcal{L}_{mms}$   & $\mathcal{L}_{cl}^{v}$ & $\mathcal{L}_{cl}^{l}$  & IoU(BG)$\uparrow$   &  IoU(FG)$\uparrow$     &   mIoU$\uparrow$   & IoU(BG)$\uparrow$   &  IoU(FG)$\uparrow$     &   mIoU$\uparrow$      \\
\midrule
S1          &\ding{55}  & \ding{55}  & \ding{55}  & 98.03  & 92.48  & 95.25 & 90.25  & 70.80  & 80.52 \\
S2          &\ding{55}  & \checkmark & \checkmark & 88.14  & 74.65  & 81.39 & 82.60  & 45.52  & 64.06 \\
S3          &\checkmark &  \ding{55} & \ding{55}  & 98.01  & 92.43  & 95.22 & 97.93  & \textbf{92.59}  & \textbf{95.26} \\
S4          &\ding{55}  & \ding{55}  & \checkmark & 98.12  & 92.86  & 95.49 & 88.38  & 62.29  & 75.34 \\
S5          &\ding{55}  & \checkmark & \ding{55}  & 87.86  & 74.20  & 81.03 & 96.62  & 88.63  & 92.62 \\
S6          &\checkmark & \checkmark & \checkmark & \textbf{98.34}  & \textbf{93.72}  & \textbf{96.03} & \textbf{97.94}  & 92.20  & 95.07 \\
\bottomrule
\end{tabular}
\vspace{-0.2cm}
\label{tab:ablation}
\end{table*}

\subsection{Segmentation}
\label{sec:seg}
As illustrated in \cref{fig:segmentation}, both patch-guided and text-guided variants of our {\M} exhibit superior object segmentation when compared to DFF-DINO, DFF-LSeg, and N3F. Furthermore, our {\M} achieves comparable performance with NeRF-SOS.
The quantitative results presented in \cref{tab:segmentation} corroborate the findings of the qualitative comparison, providing further evidence of our approach's superior performance in scene understanding and segmentation.

\begin{figure}[t]
  \centering
   \includegraphics[width=0.9\linewidth]{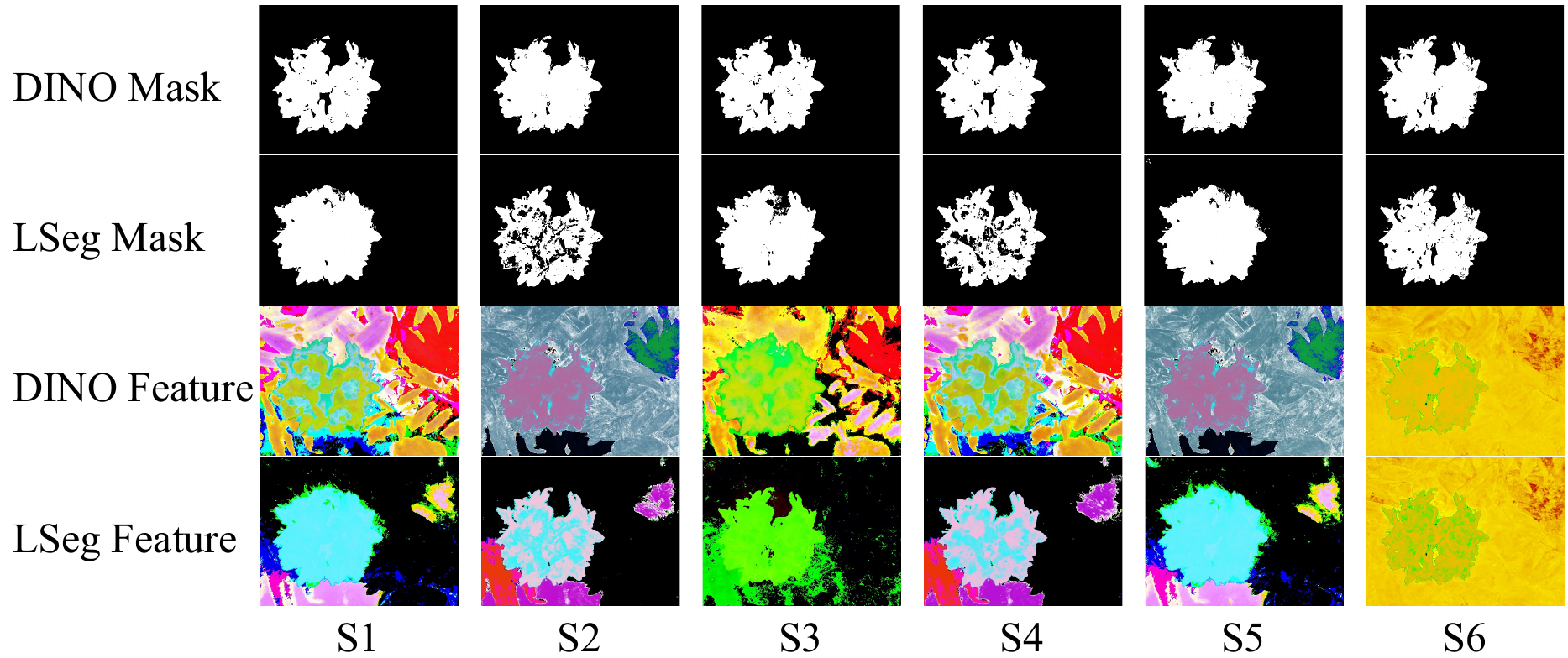}
   \caption{Ablation depicting segmentation results using DINO feature field (DINO mask) and LSeg feature field (LSeg mask), respectively. The last two rows show the corresponding features.}
   \label{fig:feature}
\end{figure}

\begin{figure}[t]
    \centering
    \includegraphics[width=\linewidth]{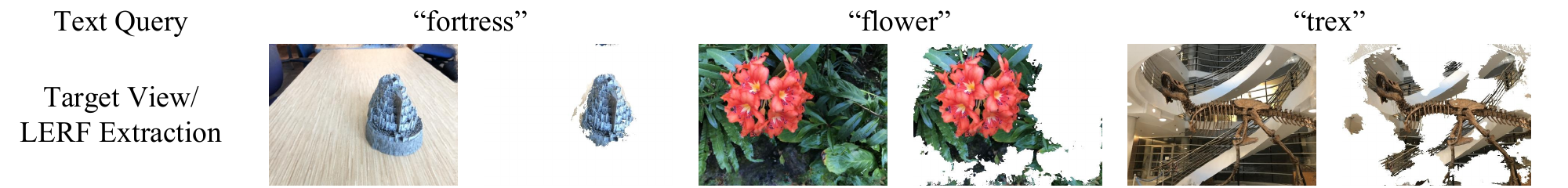}
    \vspace{-0.3cm}
    \caption{Text-based extraction for LERF.}
    \vspace{-0.6cm}
    \label{fig:lerf}
\end{figure}

\subsection{Localization}
\label{sec:localization}
LERF~\cite{lerf2023kerr} also employs a feature distillation strategy for 3D scene understanding by utilizing DINO and CLIP simultaneously. However, it ignores the relationship between visual features and language features. We compare our {\M} with LERF in the 3D localization task (\cref{fig:localization}). We see that our {\M} gives clearer and more consistent relevancy overall.  For the `horns', our model also marks the trex and floor as relevant, but is better than LERF which cannot identify the horns at all.  Overall, {\M} excels in localizing the most relevant parts with clearer boundaries, demonstrating the effectiveness of our explicit multi-modal feature fields connections. The LERF usually struggles to delineate clear boundaries accurately based on the text query (\cref{fig:lerf}).

\subsection{Ablation Study}
\label{sec:ablation}
We conduct experiments with various settings for our multi-modal similarity loss $\mathcal{L}_{mms}$ and patch-based joint contrastive loss $\mathcal{L}_{cl}$. 
We present the results in \cref{tab:ablation}, where S1 represents the multi-modal feature fields without any explicit connections, S2 denotes optimization of multi-modal feature fields with joint contrastive learning, S3 represents optimization with multi-modal similarity loss, S4 and S5 denote the application of contrastive learning for the visual feature field and language feature field separately, and S6 indicates simultaneous use of multi-modal similarity loss and joint contrastive learning.
We also show the segmentation mask and feature maps for the multi-modal feature fields for the different settings in \Cref{fig:feature}.

\noindent \textbf{Multi-Modal Similarity.}
Comparing S1 and S3 we see significant improvement in segmentation for the language feature field with the multi-modal similarity loss (\lossmms).
Between the two conditions, there is a considerable increase in IoU for LSeg in \cref{tab:ablation} and much clearer mask boundaries for LSeg Mask in \cref{fig:feature}.
Note that in the base setting (S1), the LSeg branch is conceptually similar to DFF-LSeg.
This ablation shows our {\M} model improves over DFF-LSeg (see \cref{fig:text}) by using the multi-modal similarity loss to pull together the visual and language features.

\noindent \textbf{Joint Contrastive Learning.}
For the patch-based contrastive losses, adding each helps to improve the visual and language features with clear improvements over S1 for LSeg on S4 (\lossclv) and DINO on S5 (\lossclv).
However, the straightforward combination of visual and language feature fields leads to performance degradation (S2) due to conflict in optimization for multi-modal feature volumes. By integrating multi-modal similarity loss (\lossmms) with the patch-base losses in S6, our method improves both DINO and LSeg feature branches, highlighting the importance of joint contrastive learning for multi-modal feature fields.

\noindent \textbf{Feature Visualization.}
\Cref{fig:feature} shows the clearer and more details masks we obtain with all the losses (S6) compared with the other settings. 
Although the 3D scene can be decomposed after feature distillation (S1), it fails to define the accurate boundary of a single object. 
With our multi-modal similarity loss \lossmms and joint contrastive learning scheme \losscl, the multi-modal feature fields have a consistent distribution, benefiting the decomposition for both patch-based and text-based queries.


\section{Conclusion}
\label{sec:conclusion}
We presented {\M}, a method that advances the capabilities of 3D scene understanding using 2D multi-modal features. By matching the visual and language features, our multi-modal feature fields enable patch-guided and text-guided 3D decomposition with a single NeRF model. We explored two key connection strategies: the multi-modal similarity constraint and the joint contrastive learning scheme.
Our experiments on real-world scenes demonstrate the superiority of {\M} compared to existing methods. 
Our approach handles multi-modal queries and enhances 3D scene decomposition, enabling applications in computer vision, graphics, augmented reality, and beyond.

\noindent \textbf{Limitations.} 
One limitation pertains to the inherent noise observed in density-based representations. Exploring alternative representations, such as Signed Distance Functions (SDF), might be a fruitful avenue to address this issue.
Another limitation is the absence of 3D inpainting, meaning our model may struggle with scenes where portions of the 3D environment are missing or obscured.
Addressing these limitations can lead to more robust and versatile 3D scene understanding and editing systems.
\appendix
\setcounter{page}{1}


\section{Computational Resource}
\label{sec:computation}
In our project, we compare our {\M} with DFF~\cite{dffs2022kobayashi}, N3F~\cite{n3f2022tschernezki}, NeRF-SOS~\cite{nerfsos2022fan}, and LERF~\cite{lerf2023kerr}. Our {\M} experiments are conducted on a single NVIDIA RTX A5000 GPU with 24GB memory, and DFF and N3F are executed on the same GPU.
However, NeRF-SOS has a much larger model size than ours, requiring more memory. For NeRF-SOS, the implementation is carried out on a single NVIDIA RTX A40 GPU with 48GB memory. As for LERF, it is based on NeRFStudio~\cite{nerfstudio}, which recommends a more recent CUDA version. We implement LERF on the RTX 4090 GPU with 24GB memory, which is based on CUDA 11.8. The other methods are all implemented with CUDA 11.6.

\begin{table}[h]
\centering
\caption{Computational Resource. }
\begin{tabular}{lll}
\toprule
Methods   & GPU & AVG Time  \\
\midrule
NeRF-SOS* \cite{nerfsos2022fan} & A40 (48G)        & 27.70 hours  \\
N3F \cite{n3f2022tschernezki}   & RTX A5000 (24G)  & 5.16 hours   \\
LERF \cite{lerf2023kerr}        & RTX 4090 (24G)   & 55.38 mins   \\
DFF \cite{dffs2022kobayashi}    & RTX A5000 (24G)  & 2.12 mins    \\
{\M} (Ours)                     & RTX A5000 (24G)  & 28.01 mins   \\
\bottomrule
\end{tabular}
\label{tab:time}
\end{table}

We also provide the average training time for different methods in \cref{tab:time}. The * denotes that the average time for NeRF-SOS is only calculated over \textit{\{Flower, Fortress\}} scenes. Since NeRF-SOS relies on a pretrained checkpoint for the subsequent contrastive learning, and only pretrained checkpoint models for two scenes are provided, we fine-tune NeRF-SOS on \textit{\{Flower, Fortress\}} scenes. However, even for the fine-tuning stage, NeRF-SOS demands an average of 27.70 hours. N3F requires about 5 hours for pretraining and 10 minutes for distillation. LERF requires 55.38 minutes for learning from multiple teacher models. The official code for DFF is also based on Instant NGP \cite{instantngp2022muller} (different from the setting in paper), and it only needs 2.12 minutes for optimization. Our method requires about 28.01 minutes.

\section{More Quantitative Results}
\label{sec:quantitative}

In the main paper, we compare the rendering quality with several recent methods focusing on scene decomposition and scene segmentation. NeRF-SOS~\cite{nerfsos2022fan} and Semantic-NeRF~\cite{semanticnerf2021zhi} are only evaluated over the \textit{\{Flower, Fortress\}} scenes, while the other methods are evaluated over all eight scenes. For a fair comparison, we provide an additional comparison for all methods over the \textit{\{Flower, Fortress\}} scenes in \cref{tab:rendering}. The results over the \textit{\{Flower, Fortress\}} scenes demonstrate that our method {\M} and N3F~\cite{n3f2022tschernezki} achieve the best performance for rendering.

\begin{table}[h]
\centering
\caption{Quantitative rendering results for \textit{\{Flower, Fortress\}}.}
\begin{tabular}{lccc}
\toprule
Methods      & PSNR$\uparrow$ & SSIM$\uparrow$ & LPIPS$\downarrow$ \\
\midrule
DFF-DINO \cite{dffs2022kobayashi}         & 28.03    & 0.9156    & 0.0627   \\
DFF-Lseg \cite{dffs2022kobayashi}         & 27.99    & 0.9117    & 0.0721   \\
N3F \cite{n3f2022tschernezki}             & \textbf{30.42}    & 0.9419    & 0.0379   \\
NeRF-SOS \cite{nerfsos2022fan}            & 27.85    & 0.9070    & 0.0587   \\
Semantic-NeRF  \cite{semanticnerf2021zhi} & 27.65    & 0.8039    & 0.1323   \\
LERF \cite{lerf2023kerr}                  & 14.41	 & 0.5293	 & 0.3550   \\
{\M} (Ours)                               & 30.17    & \textbf{0.9482}    & \textbf{0.0324}   \\
\bottomrule
\end{tabular}
\label{tab:rendering}
\end{table}

The LERF is built on NeRFStudio~\cite{lerf2023kerr}, and ns-eval isn't implemented inside LERF since ns-eval is for measuring metrics that aren't really relevant quantities for language. So for comparison, we have to employ ns-render, which needs the camera path and parameters (e.g. FOV) for rendering. However, we do not have the accurate FOV for each frame. Thus, the rendered images are not exactly matched with the test images (given camera paths), leading to poor quantitative results.

\begin{figure}[h]
    \centering
    \includegraphics[width=\linewidth]{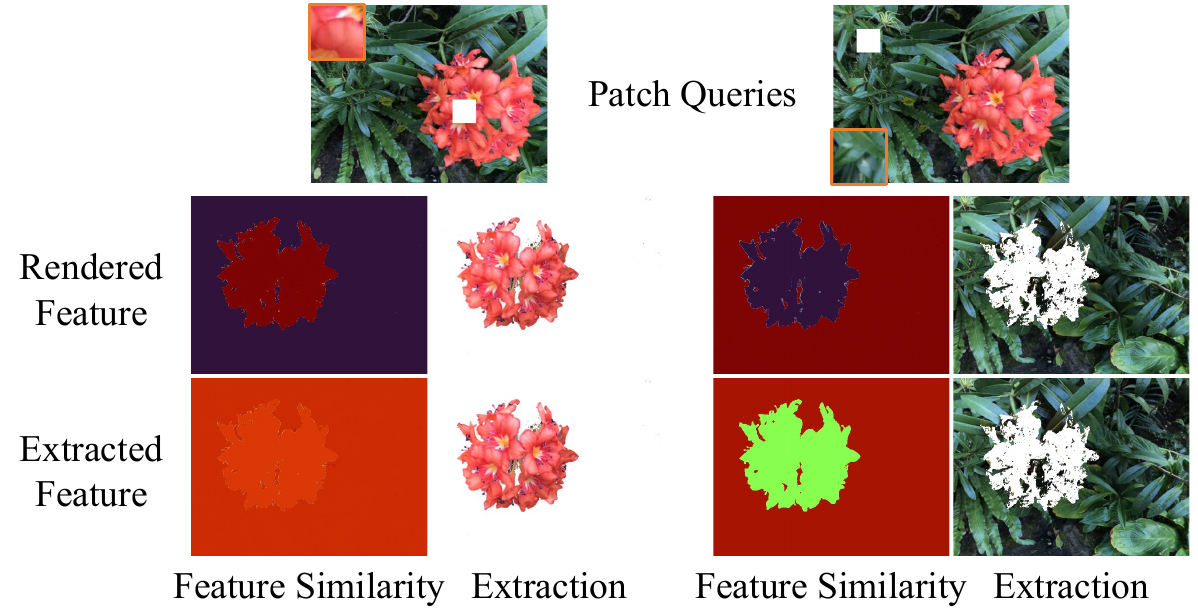}
    \caption{Matching feature using rendered features and extracted features.}
    \label{fig:2d3d}
\end{figure}

\section{More Qualitative Results}
\label{sec:qualitative}

\begin{figure}[h]
    \centering
    \includegraphics[width=\linewidth]{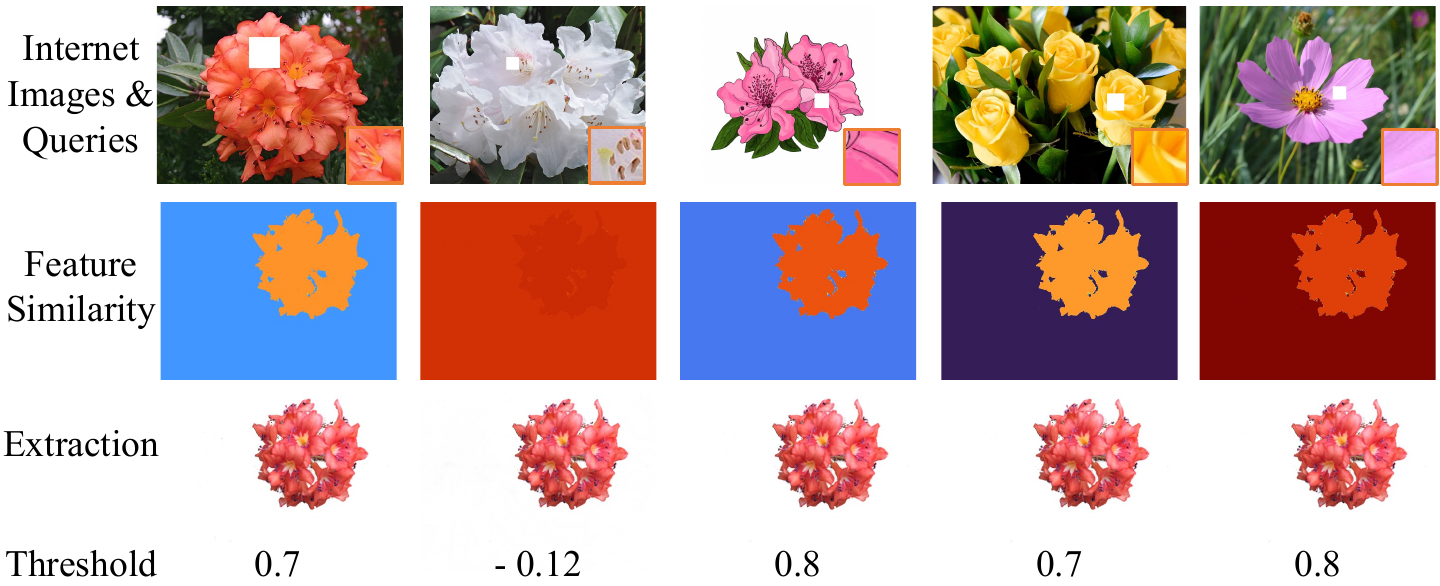}
    \caption{Editing results with patch queries from internet images.}
    \label{fig:internet}
\end{figure}

We provide more results with different queries in \cref{fig:2d3d} and \cref{fig:internet}. 
Our patch query features are derived from the averaged rendered features of specific locations in the main paper. Additionally, we conduct experiments involving feature extraction with DINO from diverse sources, encompassing images of the target scene and internet images.

When employing rendered features for feature similarity calculation (\cref{fig:2d3d}), the calculated feature similarity distinctly captures the object region corresponding to the query part. In contrast, if the features are derived from pretrained DINO with the same scene image, the feature similarity is less distinct than that achieved with rendered features. Nevertheless, the resulting edits are comparable to those based on the rendered features.

Our {\M} also supports queries from internet images (\cref{fig:internet}), whose features are also extracted with DINO. We display results of five internet images, including three different rhododendron images, one rose image, and one cosmos image. Remarkably, our method accurately identifies the relevant object through feature matching, even in cases such as the cartoon-style image of the rhododendron in the third column. 
However, we need to adjust the thresholds in some cases.

\begin{table*}[h]
\centering
\caption{Thresholds and text queries for our {\M} and DFF-LSeg. DFF-LSeg requires both positive and negative queries for each scene. Despite experimenting with thresholds up to five decimal places, the results for DFF-LSeg remain unsatisfactory.}
\begin{tabular}{lclcccll}
\toprule
\multirow{2}{*}{scenes} & \multicolumn{2}{c}{{\M} (Ours)} & \multicolumn{5}{c}{DFF-LSeg \cite{dffs2022kobayashi}}                                         \\
\cmidrule(l){2-3} \cmidrule(l){4-8}
                        & threshold            & text query & \multicolumn{3}{c}{thresholds} & positive query & negative queries     \\
\midrule
fern                    & 0.8    & fern       & 0.24400  & 0.24420  & 0.24425 & fern           & floor, wall, ceiling \\
flower                  & 0.8    & flower     & 0.38160  & 0.38170  & 0.38180 & flower         & leaves, ground       \\
leaves                  & 0.8    & leaves     & 0.50800  & 0.50805  & 0.50810 & leaves         & background           \\
trex                    & 0.8    & trex       & 0.22550  & 0.22600  & 0.22650 & trex           & floor, wall, ceiling \\
\bottomrule
\end{tabular}
\label{tab:text-thr}
\end{table*}

\section{Implementation Details}
\label{sec:implementation}

\subsection{Thresholds Selecting}

For our {\M}, we use a fixed threshold of 0.8 for all queries and all scenes. 
In contrast, both DFF and N3F require manual adjustment of thresholds to find the most suitable thresholds for each query and scene. The thresholds for text queries and patch queries are presented in \cref{tab:text-thr} and \cref{tab:patch-thr}.

For text-based editing (\cref{tab:text-thr}), DFF-LSeg~\cite{dffs2022kobayashi} additionally requires both positive and negative queries for each scene. Despite experimenting with thresholds up to five decimal places, the results for DFF-LSeg remain unsatisfactory. Our {\M} only needs the text query for the target object and gets much better editing results with a fixed threshold.

\begin{table}[h]
\centering
\caption{Thresholds for our {\M}, DFF-Lseg, and N3F. }
\begin{tabular}{lccc}
\toprule
scenes   & {\M} (Ours) & DFF-DINO \cite{dffs2022kobayashi} & N3F \cite{n3f2022tschernezki} \\
\midrule
fern     & 0.8    & 0.4      & 1.2 \\
flower   & 0.8    & 0.6      & 1.2 \\
fortress & 0.8    & 0.5      & 0.7 \\
orchids  & 0.8    & 0.6      & 1.2 \\
\bottomrule
\end{tabular}
\label{tab:patch-thr}
\end{table}
For patch-based editing (\cref{tab:patch-thr}), the thresholds are much easier to select than text-based editing, since the DINO contains much clearer semantic features than LSeg. However, we still need to try different thresholds for different scenes in DFF-DINO~\cite{dffs2022kobayashi} and N3F~\cite{n3f2022tschernezki}. 

\subsection{Segmentation}
For the segmentation task, we mostly follow NeRF-SOS~\cite{nerfsos2022fan}'s segmentation approach, which generates segmentation results based on K-means clustering on the segmentation logits. Compared with NerF-SOS, we don't have a segmentation branch that outputs the segmentation logits specifically for clustering. Instead, we use K-means clustering for the multi-modal feature fields. The segmentation results for {\M}-patch and {\M}-text are based on the K-means of the visual feature field and language feature field, respectively. The number of clusters is set to 2 for all scenes to decompose the scene as foreground and background.

%
%
\bibliographystyle{splncs04}
\bibliography{main}

\end{document}